\documentclass[conference]{IEEEtran}
\IEEEoverridecommandlockouts
\usepackage{graphicx}
\usepackage{color}
\usepackage{chngpage}
\usepackage{multicol}
\usepackage{array,multirow}
\usepackage{amssymb}
\usepackage{amsmath}
\usepackage{bm}
\usepackage{subcaption}
\usepackage{tabularx,booktabs}
\usepackage{hyperref}

\newcolumntype{Y}{>{\centering\arraybackslash}X}

\usepackage[utf8]{inputenc}

\usepackage{xcolor}
\definecolor{RoyalBlue}{RGB}{65, 105, 225}

\newcolumntype{?}{!{\vrule width 1pt}}

\def\BibTeX{{\rm B\kern-.05em{\sc i\kern-.025em b}\kern-.08em
    T\kern-.1667em\lower.7ex\hbox{E}\kern-.125emX}}
\begin{document}

\title{Joint 2D-3D Breast Cancer Classification \\
% \thanks{\todo{Identify applicable funding agency here}}
% \thanks{Code is available at: \url{https://github.com/XXXX/XXXX}}
\thanks{This study is sponsored by Grant No. IRG 16-182-28 from the American Cancer Society and Grant No. IIS-1553116 from the National Science Foundation.}
% \thanks{Code available at \href{https://github.com/GB-TonyLiang/joint_2d-3d_mammo}{https://github.com/GB-TonyLiang/joint\_2d-3d\_mammo}}
}

\author{\textbf{Gongbo~Liang\textsuperscript{1}*,
        Xiaoqin~Wang\textsuperscript{2}, 
		Yu~Zhang\textsuperscript{1},
		Xin~Xing\textsuperscript{1},
		Hunter~Blanton\textsuperscript{1},
		Tawfiq~Salem\textsuperscript{3},
		Nathan~Jacobs\textsuperscript{1}} \\ [1ex]
		$1$ Department of Computer Science, University of Kentucky, Lexington, KY, USA \\
		$2$ Department of Radiology, University of Kentucky, Lexington, KY, USA \\
		$3$  Department of Computer Science and Software Engineering, Miami University, Oxford, Ohio, USA\\
		Email: liang@cs.uky.edu*}
		
% \author{\textbf{Author}}

\maketitle              

\begin{abstract}
Breast cancer is the malignant tumor that causes the highest number of cancer deaths in females. Digital mammograms (DM or 2D mammogram) and digital breast tomosynthesis (DBT or 3D mammogram) are the two types of mammography imagery that are used in clinical practice for breast cancer detection and diagnosis. Radiologists usually read both imaging modalities in combination; however, existing computer-aided diagnosis tools are designed using only one imaging modality. Inspired by clinical practice, we propose an innovative convolutional neural network (CNN) architecture for breast cancer classification, which uses both 2D and 3D mammograms, simultaneously. Our experiment shows that the proposed method significantly improves the performance of breast cancer classification. By assembling three CNN classifiers, the proposed model achieves $0.97$ AUC, which is $34.72\%$ higher than the methods using only one imaging modality.

\end{abstract}


\section{Introduction}
Breast cancer is the leading cause of cancer death in over $100$ countries~\cite{bray2018global,siegel2019cancer}. % and the most frequently diagnosed cancer in $154$ out of $195$ countries in the world~\cite{bray2018global,siegel2019cancer}. 
Mammography is the only image screening tool that has been proven to reduce breast cancer mortality~\cite{henrot2014breast}. Digital mammography (DM or 2D mammography) and digital breast tomosynthesis (DBT or 3D mammography) are the two types of mammograms that are used in clinical practice %(Figure~\ref{fig:dm_dbt})
~\cite{american2019breast}. 
Radiologists usually read both imaging modalities in combination, often looking for changes from slice-to-slice in DBT and comparing that with structures in DM~\cite{friedewald2014breast}. However, interpreting mammograms is a challenging task, requiring many years of professional training. It is also time-consuming and therefore an expensive process~\cite{giger2013breast}. This is especially problematic given the worldwide shortage of specialized breast radiologists~\cite{wing2009workforce}. 

Deep learning has demonstrated revolutionary potential in medical imaging analysis~\cite{esteva2017dermatologist, mihail2019automatic, liang2018ganai}. Given the need for highly efficient and accurate mammogram analysis, numerous deep learning-based computer-aided diagnosis (CAD) models have been developed~\cite{ribli2018detecting,zhang2018classification,mendel2019transfer}. However, the existing models typically focus on using either DM or DBT. 

% \begin{figure}[!tb]
% \centering
% \includegraphics[width=0.45\textwidth]{figures/dm_dbt.png}
% \caption{Example of a positive 2D digital mammogram in craniocaudal (CC) view (left) and the four slices from the corresponding digital breast tomosynthesis (right).}
% 	\label{fig:dm_dbt}
% \end{figure}

Inspired by clinical practice, we propose a novel breast cancer classification approach using convolutional neural networks (CNN) combined with ensemble strategy. The proposed network simultaneously reads DM and DBT as what radiologists would do in their daily practice. One key challenge of this work is how to use DBT effectively.  The data size of DBT is large and with varying depths (on average, each DBT has $1024\times 1024\times 82$ voxels in this study). Training a 3D CNN model for such large data is extremely costly in terms of computation and memory, and may potentially lead to overfitting. We innovatively extract a fixed-size slice representation for each DBT, which captures the changes between DBT slices, and use a 2D CNN for classification. From our experiment, the proposed method has improved the performance significantly. In summary, the proposed method has the following advantages.
\begin{itemize}
\item To our best knowledge, this is the first model using whole DM and DBT simultaneously.
\item We innovatively extract a fixed-size representation for each DBT. The extracted representation captures the changes between different slices of the same DBT.
\item We use a real-world clinical dataset in this study. To our best knowledge, this is the largest breast cancer dataset that contains paired DM and DBT.
\item Our method only requires image-level labels for training and significantly improves performance compared to other approaches trained similarly.% (from $0.72$-$0.87$~\cite{zhang2018classification} to $0.97$ AUC). 
\end{itemize}

\begin{figure*}[!ht]
\centering
\includegraphics[width=0.925\textwidth]{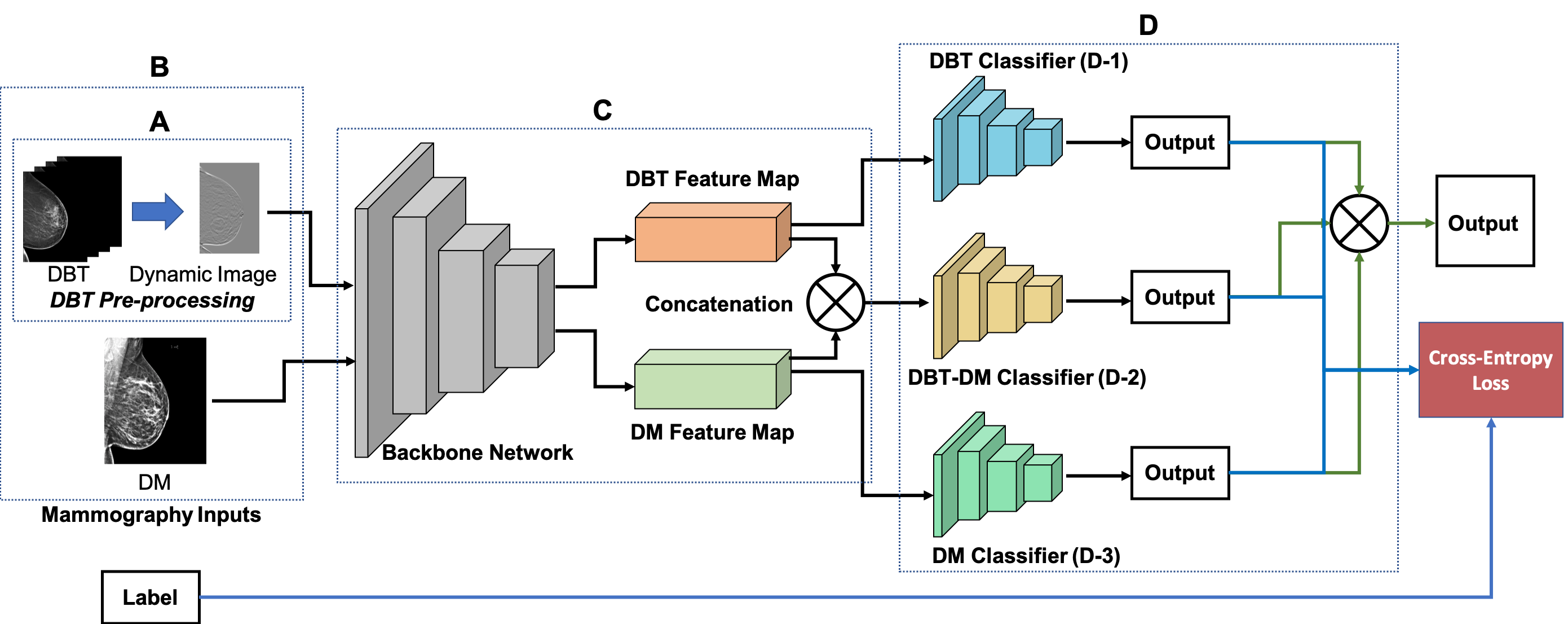}
\caption{Overview of the proposed model. A) DBT data pre-processing. B) DM/DBT pairs as the input of the model. C) Feature map extraction using a backbone network. D) Ensemble outputs of different CNN classifiers for testing. Blue lines, model training stage; Green lines, model testing stage; Black lines, shared by the training and testing stage; DBT, digital breast tomosynthesis; DM, digital mammogram.}
	\label{fig:overview}
\end{figure*}

\section{Background}
% \subsection{Deep Learning Based Computer-Aided Diagnosis Tools}
\subsection{Existing Deep Learning Models}
Ribli et al. used an r-CNN-based~\cite{ren2015faster} approach to classify the 2D mammograms and that achieved $0.95$ AUC (area under the receiver operating characteristic curve) for breast tumor classification~\cite{ribli2018detecting}. This work won $2^{nd}$ place at the Digital Mammography DREAM Challenge~\cite{dream2016}. Shen et al.~\cite{shen2017end} designed a fully convolutional network for mammogram classification, which achieved $0.94$ AUC. %Both models were developed on an outdated scanned film mammography dataset, the DDSM dataset~\cite{lee2016curated}, which was established in the $1990s$.
Though the performances for these two models are impressive, both works were trained using bounding boxes (BBs), which are usually not available on clinical data due to the high obtaining cost for medical images. More importantly, these methods only designed for 2D mammograms. None of them works on DBT.

Mendel et al.~\cite{mendel2019transfer} proposed a model using a pre-trained VGG19~\cite{simonyan2014very} network as the feature extractor and using support vector machine (SVM) %~\cite{cortes1995support} 
as the classifier to separately evaluate breast lesions in DM and DBT. They reported $0.81$ and $0.89$ AUC on DM and DBT, respectively. However, the proposed work has several major limitations. For instance, a keyframe of each DBT needs to be selected by a trained radiologist during the data pre-processing step. This human involvement does not only increase the cost of using the method but also introduces bias into the proposed model. More importantly, this human pre-selection step treats DBT as an additional DM, which omits using the most important information of DBT—the slice-to-slice changes.  In addition, this method also requires BBs, which are usually not available to clinical practice.%, which are often used by radiologists for breast cancer detection and diagnose.%Secondly, the clinical dataset used in the study is small, which only contains $78$ lesions from $76$ patients. In addition, this method also requires BBs, which are usually not available to clinical practice. Most importantly, such a method omits using the most important informant of DBT—the slice-to-slice changes, which are often used by radiologists for breast cancer detection and diagnose.

Zhang et al.~\cite{zhang2018classification} proposed an end-to-end breast cancer classification method using AlexNet~\cite{krizhevsky2012imagenet} as the backbone. Their method does not require BBs for training, but it still needs to use two different models to evaluate DM and DBT separately. Though their model has some advantages over the previous ones, the model performs poorly on DBT due to the high computational cost of 3D CNN model. They only reported a $0.66$ AUC on normal vs. malignant classification. %In addition, the clinical dataset used in their study is imbalanced. It has over $800$ negative patients and only $125$ positive patients (only about $13.5\%$ of the total data are positive).

\subsection{CNN Model for Volumetric Data}
Two types of 3D CNNs are widely used for volumetric data classification. One is the fully 3D CNN architecture, such as I3D~\cite{carreira2017quo} and 3D-ResNet~\cite{hara2018can}. The second is to use 2D CNN models in a 3D way, such as~\cite{zhang2018classification}. Even though the two approaches work differently, they both suffer from the same limitations. For instance, volumetric data usually have much more extensive data size than a regular image. The average size of ImageNet data is $469\times387$ pixels, but the average size of DBT used in this study is $1024\times1024\times82$ voxels. Training a 3D CNN model on such a large data size is extremely computationally costly and may potentially lead to overfitting. To reduce the negative effect,~\cite{zhang2018classification} takes only $30$ slices of each DBT as the input. However, by doing this, either a pre-selection step is needed, or we could only hope the slices we decide to feed into the model will represent the whole volumetric data sufficiently. Neither of the scenarios is optimal. Thus, directly training a 3D CNN model for DBT may not be a good option.

\section{Architecture Overview}
We propose a novel CNN ensemble method for breast tumor classification. The proposed approach consists of three main components: 1) DBT pre-processing approach (Figure~\ref{fig:overview}A), 2) DBT and DM feature extraction and feature map concatenation (Figure~\ref{fig:overview}C), and 3) multiple classifiers and ensemble outputs of each classifier (Figure~\ref{fig:overview}D). 

\subsection{DBT Pre-processing}
\begin{figure}[!tb]
  \centering
    \includegraphics[width=0.415\textwidth]{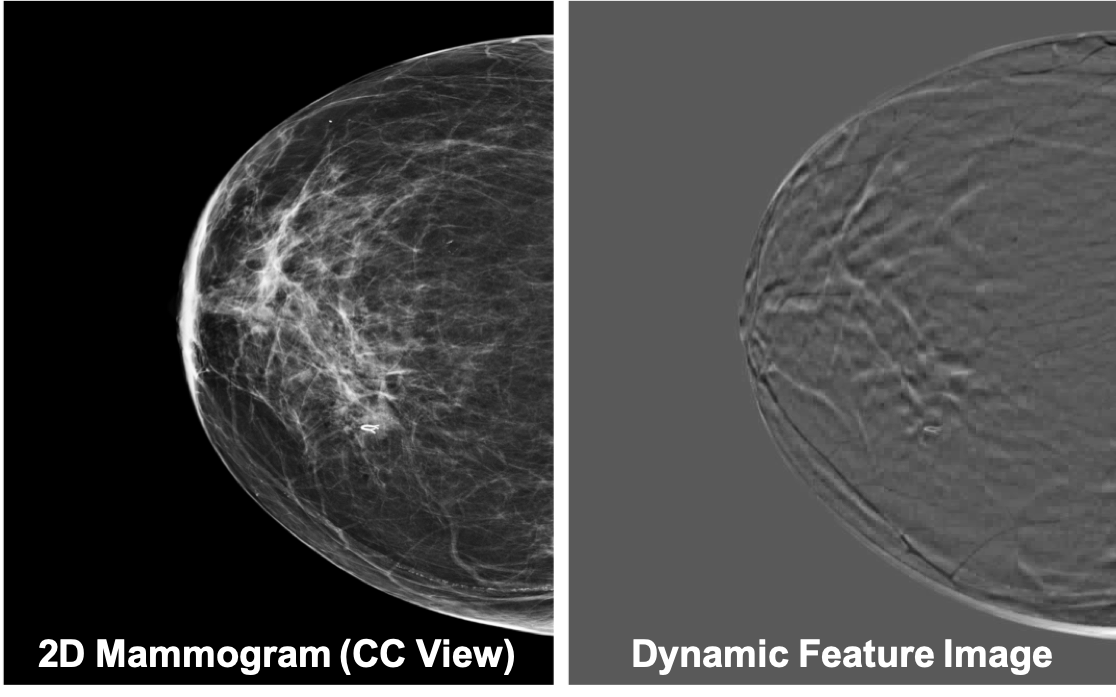}
	\caption{Example of DM and the corresponding dynamic feature images.}
    \label{fig:mammo_dynamic}
\end{figure}

In non-medical domain, a popular method to represent a series of images is to apply a temporal pooling operator to the features extracted at individual images, for instance, temporal templates~\cite{bobick2001recognition}, ranking functions~\cite{fernando2015modeling} and sub-videos~\cite{hoai2014improving}, as well as other traditional pooling operators~\cite{ryoo2015pooled}. We adopt the idea of temporal pooling operator to the medical imaging domain. Inspired by Bilen et al., we applied RankSVM~\cite{bilen2016dynamic} directly on DBT data to extract a fixed, one-slice representation of each DBT. Since the extracted fixed representation keeps the dynamic features (i.e., the slice-to-slice changes) of DBT, we call it \textit{dynamic feature image}. See Figure~\ref{fig:mammo_dynamic} for an example.

One dynamic feature image is a single RGB image, which captures the slice-to-slice changes of a DBT. A ranking function is used to obtain the dynamic feature image for a series of slices $I_1$,...,$I_T$, temporally. More specifically, let $\psi(I_t) \in \mathbb{R}^d$ be the feature vector extracted from each individual slices $I_t$ in the series. Let $V_t=\frac{1}{t}\sum_{\tau=1}^t \psi(I_\tau)$ be the average time of these features up to time $t$. The ranking function associates to each time $t$ a score $S(t|\bm{d})=\langle\bm{d}, V_t\rangle$, where $\bm{d} \in \mathbb{R}^d$ is a vector of parameters. The function parameters $\bm{d}$ are learned so that the scores reflect the rank of the slices in the series. Therefore, later times are associated with larger scores, i.e. $q\succ t \Rightarrow S(q|\bm{d}) > S(t|\bm{d})$. 
Learning $\bm{d}$ is posed as a convex optimization problem using the RankSVM function:
\begin{equation}
  \begin{array}{l}
      \bm{d}^* = \rho(I_1,...,I_T;\psi) = \operatorname*{argmin}_d E(\bm{d}), \\
          \begin{aligned}
            E(\bm{d})=&\frac{\lambda}{2}||\bm{d}|| ^2+\frac{2}{T(T-1)}\times \\
            &\sum_{q>t}\max\{0,1-S(q|\bm{d})+S(t|\bm{d})\}.
          \end{aligned}
  \end{array}
\end{equation}
The first term in the objective function is a quadratic regularizer used in SVM. The second term is a hinge-loss that counts how many pairs $q\succ t$ are incorrectly ranked by the scoring function. The optimizer to the RankSVM is written as a function $p(I_1, ..., I_t ; \psi)$ that maps a series of $T$ slices to a single vector $\bm{d^*}$. Since this vector contains enough information to rank all the slices in the series, it aggregates information from all of them and can be used as a descriptor of a series of slices. The process of constructing $\bm{d^*}$ is known as rank pooling~\cite{fernando2016rank}, which can be applied to DBT directly.

\subsection{CNN Architectures}
The proposed network contains two kinds of CNNs: the backbone CNN feature extracting network (feature extractor) and the shallow CNN classifier (classifier).  

\subsubsection{CNN Feature Extractor}
The feature extractor is a fully convolutional network (FCN), which takes a $W\times H\times K$ image as input and output of a $W'\times H'\times K'$ feature map. We use the common CNN classification architecture to build the feature extractor
by pre-training it on ImageNet~\cite{deng2009imagenet} dataset. After the model is well-trained, the fully connected (FC) layers of the model are removed. The pooling layer between the first FC layer and the last convolution (Conv) layer is also removed, if applicable. We use the output of the last convolutional layer of the model as the extracted feature map. All the parameters are frozen during the feature extracting step. 

\subsubsection{CNN Classifier}
\begin{figure}[!tb]
	\centering
	\includegraphics[width=0.45\textwidth]{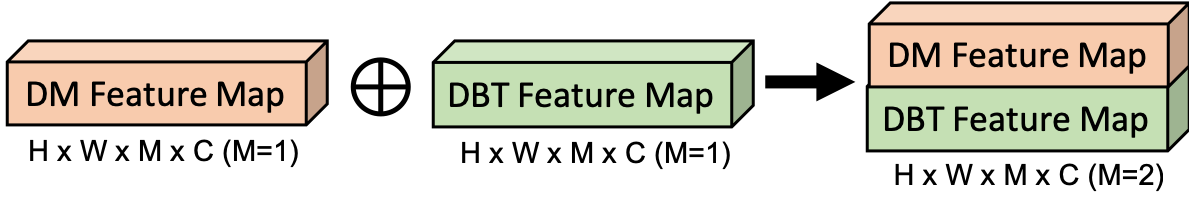}
	\caption{Concatenate features on $modality$ dimension. H=height, W=width, M=modality, C=channel.}
	\label{fig:concatenate_feature}
\end{figure}

There are three CNN classifiers with two different architectures included in the proposed model. The DBT Classifier and DM Classifier (Figure~\ref{fig:overview}D-1 and~\ref{fig:overview}D-3) are used for DBT feature map classification and DM feature map classification, respectively. These two classifiers share the same architecture but with different weights, which was implemented as a 2D Conv layer followed by two FC layers.  The DM-DBT Classifier (Figure~\ref{fig:overview}D-2) simultaneously evaluates the DM and DBT by taking the feature maps of the two imaging modality in combination. Since we concatenated the two feature maps on the $modality$ dimension, the dimensionality of the feature map is increased by $1$ (see Figure~\ref{fig:concatenate_feature}). We replace the 2D Conv layer in the other classifiers with a 3D Conv layer. The 3D Conv kernels are applied on the $height$, $width$, and $modality$ dimensions. Both of the 2D and 3D Conv layer included convolution, batch normalization, leaky ReLU, and max pooling. The batch size is $32$. Max pooling has a $2\times2$ or $2\times2\times1$ receptive field with stride $1$ for 2D or 3D Conv layer, respectively. Cross-entropy loss is used in training. Adam optimizer with a learning rate of $0.0001$ is used as the optimizer. Dropout with a rate of $0.5$ is applied to the FC layers. See Table~\ref{table:cnn_classifier} for Classifier architecture detail.

\begin{table}[!tb]
    \begin{adjustwidth}{-1in}{-1in}% adjust the L and R margins by 1 inch
    \caption{Detail of CNN Classifiers.}
	\centering
	\setlength\tabcolsep{2pt}
	\begin{tabular}{| c | c | c | c | c | c | c | c |} 
		\hline
		\textbf{Classifier} &\textbf{Input Shape} &  \textbf{Conv Layer}  &  \textbf{Conv Type} & \textbf{Pooling} &  \textbf{FC1} &  \textbf{FC2}\\[0.5ex] 
        \hline
        DBT or DM &$w\times h\times c$ & $c$ @ $1\times1$ & 2D Conv & $2\times2$ & $256$ & $128$ \\ 
        \hline
        DM-DBT &$w\times h\times2\times c$ & $c$ @ $1\times1\times2$ & 3D Conv & $2\times2\times1$ & $256$ & $128$ \\  
        [0.5ex] 
		\hline           
	\end{tabular}
    \label{table:cnn_classifier}
	\end{adjustwidth}
\end{table}

\subsection{Classifier Ensemble}
We propose to use the ensemble learning strategy to improve both the model performance and prediction confidence. In order to keep our method intuitive and straightforward, we use the majority voting strategy~\cite{james1998majority} in this study.

Suppose we have $K$ classifiers, the majority voting can be computed as:  
\begin{equation}
  \begin{array}{l}
      C(X)=\operatorname*{arg\max}_i \sum_{j=1}^K w_jI(h_j(X)=i),
  \end{array}
\end{equation}
where $h_i$ is the classifier, $w_i$ is the weights that sum to 1, and $I(\cdot)$ is an indicator function.

\section{Evaluation}
\subsection{Dataset}
A private clinical dataset is used in this study. All the DM and DBT data were retrospectively collected from patients seen at the University of Kentucky Medical Center from Jan $2014$ to Dec $2017$. The dataset contains $415$ benign patients and $709$ malignant patients. Each patient was reviewed by practicing breast radiologists. Both the benign and malignant cases were proved with a biopsy. All patients had both DM and DBT in either craniocaudal (CC) or mediolateral oblique (MLO) view or both views. Approximately $1400$ paired DM/DBT data were included. To our best knowledge, this is the largest paired DM/DBT breast cancer dataset.

The DM was provided in 12-bit DICOM format at $3328\times4096$ resolution. The DBT was provided in 8-bit AVI format with a resolution of $1024\times1024$. All the frames of each DBT data was saved to a set of 8-bit JPEG images before generating the dynamic feature images. Both the DM and dynamic feature images were down sampled to $832\times832$. Data augmentation was also applied to each of the mammography images and dynamic feature image through a combination of reflection and rotation. Each original image was flipped horizontally and rotated by each of $90$, $180$, and $270$ degrees. In total, $6875$ paired DM/DBT data were used in this study.

The dataset was randomly partitioned into training and testing datasets with a $4:1$ ratio on the patient-level. All the images of the same patient will be in either the training set or the test set. The benign and malignant ratio was maintained in both training and testing sets. To minimize the imbalance effect (low benign to malignant ratio), we balanced each mini-batch during training.

\subsection{Implementation and Evaluation Metrics}
Four popular CNN networks were used as the backbone feature extractor in this study,  namely AlexNet~\cite{krizhevsky2012imagenet}, ResNet~\cite{he2016deep}, DenseNet~\cite{huang2017densely}, and SqueezeNet~\cite{iandola2016squeezenet}. The model was implemented in Pytorch~\cite{paszke2017automatic}, and trained with balanced mini-batches for $100$ epochs on a Linux computer server with eight Nvidia GTX $1080$ GPU cards. %The whole training process, including the feature extraction and the CNN classifier training, can be done within several hours. 

The classification accuracy (ACC), area under the receiver operating characteristic curve (AUC), precision (Prec), recall (Reca), F1 score, average precision (AP), and average correct predict confident (AC) were used as the evaluation metrics in this study.

\subsection{Baseline Model and Ensemble Approach}
We use the 2D-T3-Alex and 3D-T2-Alex models from~\cite{zhang2018classification} as the baseline model for DM and DBT, respectively. The 2D-T3-Alex model is a transfer learning 2D CNN model, which uses pre-trained AlexNet to extract features. The 3D-T2-Alex model is a 3D CNN model, which firstly uses the regular AlexNet model to extract feature maps of every slice in a DBT. Then, $K$ feature maps of each DBT are fed into a one-Conv-layer 3D CNN model for classification. $K = 30$ was chosen in their paper. 

Our experiment shows the proposed model significantly improves the performance. By only using DBT data (i.e., the dynamic feature images), the performance can be improved from $0.72$ AUC to $0.89$ AUC ($23.61\%$ increasing). When using DM and DBT in combination, a single model can achieve $0.95$ AUC. After assembling the three classifiers (DM Classifier, DBT Classifier, and DM-DBT Classifier, which uses DM only, DBT only, and DM and DBT data, respectively), the proposed model can further improve the performance to $0.97$ AUC (Table~\ref{table:baseline}).

Table~\ref{table:ensemble} lists the ensemble result of all different backbone networks. The performance is consistency among the four different feature extractors, which indicates the proposed method is not limited to any specific architecture.

% \begin{table}[!tb]
%   \caption{Ensemble results for different backbone networks.}
%   \centering
% %   \setlength\tabcolsep{15pt}
%   \begin{tabular}{|c||c|c|c|}
%     \hline
%     \textbf{Model} & \textbf{Input Data} & \textbf{Backbone Network} & \textbf{AUC} \\
%     \hline
%       $Baseline-2D$ & DM only & AlexNet & $0.87$  \\
%       $Baseline-3D$ & DBT only & AlexNet & $0.72$ \\
%     \hline
%       $Ours_{Ensemble}$ & DM \& DBT & AlexNet & $0.97$\\
%       $Ours_{Ensemble}$ & DM \& DBT & ResNet & $0.96$\\
%       $Ours_{Ensemble}$ & DM \& DBT & DenseNet & $0.97$\\
%       $Ours_{Ensemble}$ & DM \& DBT & SqueezeNet & $0.97$\\
%     \hline
%   \end{tabular}
%   \label{table:baseline}
% \end{table} 

\begin{table}[!tb]
  \caption{Ensemble results for different backbone networks.}
  \centering
  \begin{tabular}{|c||c|c|c|}
    \hline
    \textbf{Model} & \textbf{Input Data} & \textbf{Backbone Network} & \textbf{AUC} \\
    \hline
       $2D-T3-Alex$ & DM only & AlexNet & $0.87$  \\
       $3D-T2-Alex$ & DBT only & AlexNet & $0.72$ \\
    \hline
       $Ours_{DBT}$ & DBT only & AlexNet & $0.89$ \\
       $Ours_{DM-DBT}$ & DM \& DBT & AlexNet & $0.95$  \\
       $Ours_{Ensemble}$ & DM \& DBT & AlexNet & $\bf{0.97}$\\
    \hline
  \end{tabular}
   \label{table:baseline}
\end{table} 

\begin{table}[!tb]
  \caption{Comparing with baseline model.}
  \centering
  \begin{tabular}{|c||c|c|}
    \hline
    \textbf{Backbone Network} &\textbf{Input Data} &  \textbf{AUC} \\
    \hline
       AlexNet & DM \& DBT & $0.97$  \\
       ResNet & DM \& DBT & $0.96$  \\
       DenseNet & DM \& DBT & $0.97$  \\
       SqueezeNet & DM \& DBT & $0.97$  \\
    \hline
  \end{tabular}
   \label{table:ensemble}
\end{table}

\subsection{Single Modality vs.\ Multiple Modalities}
In this section, we evaluate the model performance using a single imaging modality vs. multiple imaging modalities. More specifically, we are comparing the performance of DM Classifier, DBT Classifier, and DM-DBT Classifier. Four different backbone networks were used. In total, 12 models were trained and compared in this experiment. 

Table~\ref{table:results} reveals when using multiple imaging modalities together, the model performance is significantly better. The DM-DBT Classifier achieves a $0.95$ AUC on average. However, the save metric for DM Classifier and DBT Classifier is $0.88$ and $0.89$, respectively. The table also shows when using DBT data, the model prediction confidence can be improved, especially when using DM and DBT in combination. On average, the prediction confidence of DM Classifier is $0.83$, the same metric of DBT Classifier and DM-DBT Classifier is $0.89$ and $0.93$, respectively. As in the previous section, the performance of all four different backbone networks is consistent. They all achieved a similar result, except the average prediction confidence of single modality classifiers (i.e., DM Classifier and DBT Classifier). Among the four backbone networks, the DenseNet performance is slightly better than others, which achieves the highest scores of $17$ out of $21$ different metrics for different classifiers. 

% Figure~\ref{fig:result} shows the AUC curve of the different models. The green line indicates the AUC performance of DM-DBT Classifier, the orange line is for the DBT Classifier, the blue line is for the DM Classifier, and the dashed line indicates a random performance ($0.5$ AUC). 

\begin{table*}[!ht]
	\centering
	\caption{Evaluation results of models trained with a single modality vs.\ models trained with the multiple modalities.}
    \setlength\tabcolsep{2.5pt}
    \begin{tabularx}{\textwidth}{ |c|c|c|c|c|c|c|c?c|c|c|c|c|c|c?c|c|c|c|c|c|c|} 
    \cline{1-22}
    % \multirow{2}{*}{\textbf{Feature}} &
    \textbf{Backbone} & \multicolumn{7}{c|}{\textbf{DM Classifier}} & \multicolumn{7}{c|}{\textbf{DBT Classifier}} & \multicolumn{7}{c|}{\textbf{DM-DBT Classifier}} \\ \cline{2-22}
    \textbf{Network} & \textbf{ACC} & \textbf{AUC} & \textbf{F1} & \textbf{Prec} & \textbf{Reca} & \textbf{AP} & \textbf{AC} & \textbf{ACC} & \textbf{AUC} & \textbf{F1} & \textbf{Prec} & \textbf{Reca} & \textbf{AP} & \textbf{AC} & \textbf{ACC} & \textbf{AUC} & \textbf{F1} & \textbf{Prec} & \textbf{Reca} & \textbf{AP} & \textbf{AC} \\ 
    \cline{1-22}
    AlexNet & $0.78$ & $0.87$ & $0.76$ & $\bf{0.87}$ & $0.75$ & $0.70$ & $0.78$ & 
    $0.81$ & $0.89$ & $0.80$ & $0.84$ & $0.76$ & $0.76$ & $0.83$ & 
    $0.90$ & $0.95$ & $0.89$ & $0.91$ & $0.87$ & $0.86$ & $0.83$\\
    
    ResNet & $0.78$ & $0.87$ & $0.79$ & $0.75$ & $\bf{0.83}$ & $0.71$ & $\bf{0.96}$ & 
    $0.79$ & $0.88$ & $0.79$ & $0.80$ & $0.78$ & $0.73$ & $\bf{0.97}$ & 
    $0.87$ & $0.94$ & $0.87$ & $0.82$ & $\bf{0.93}$ & $0.80$ & $0.96$ \\
    
    DenseNet & $\bf{0.79}$ & $\bf{0.90}$ & $\bf{0.80}$ & $0.76$ & $\bf{0.83}$ & $\bf{0.72}$ & $0.79$ & 
    $\bf{0.85}$ & $\bf{0.91}$ & $\bf{0.85}$ & $\bf{0.86}$ & $\bf{0.84}$ & $\bf{0.80}$ & $\bf{0.97}$ & 
    $\bf{0.91}$ & $\bf{0.96}$ & $\bf{0.91}$ & $\bf{0.93}$ & $0.89$ & $\bf{0.88}$ & $0.96$ \\
    
    SqueezeNet & $0.78$ & $0.88$ & $0.78$ & $0.78$ & $0.79$ & $0.71$ & $0.80$ & 
    $0.79$ & $0.89$ & $0.78$ & $0.85$ & $0.72$ & $0.75$ & $0.79$ & 
    $0.90$ & $\bf{0.96}$ & $\bf{0.91}$ & $\bf{0.93}$ & $0.88$ & $\bf{0.88}$ & $\bf{0.97}$ \\
    \cline{1-22}
    Average & $0.78$ & $0.88$ & $0.78$ & $0.79$ & $0.80$ & $0.71$ & $0.83$ & 
    $0.81$ & $0.89$ & $0.81$ & $0.84$ & $0.78$ & $0.76$ & $0.89$ &
    $0.89$ & $0.95$ & $0.90$ & $0.90$ & $0.90$ & $0.86$ & $0.93$ \\
    \cline{1-22}

    \end{tabularx}
    \label{table:results}
\end{table*}

% \begin{figure}[!tb]
%   \centering
%     \begin{subfigure}[b]{0.241\textwidth}
%         \includegraphics[width=\textwidth]{figures/alex_auc.png}
%         \caption{AlexNet Backbone Models}
%         %\label{}
%     \end{subfigure}
%     \begin{subfigure}[b]{0.241\textwidth}
%         \includegraphics[width=\textwidth]{figures/res_auc.png}
%         \caption{ResNet Backbone Models}
%         %\label{}
%     \end{subfigure}

%     \begin{subfigure}[b]{0.241\textwidth}
%         \includegraphics[width=\textwidth]{figures/dense_auc.png}
%         \caption{DenseNet Backbone Models}
%         %\label{}
%     \end{subfigure}
%     \begin{subfigure}[b]{0.241\textwidth}
%         \includegraphics[width=\textwidth]{figures/squeeze_auc.png}
%         \caption{SqueezeNet Backbone Models }
%         %\label{}
%     \end{subfigure}
    
% 	\caption{AUC curves of different models. By using 2D digital mammogram (DM) and digital breast tomosynthesis (DBT) data in combination, the model performance can be improved significantly. DM Classifier-Use only DM data for training, DBT Classifier-Use only DBT data for training, DM-DBT Classifier-Use DM and DBT data in combination for training.}
%   \label{fig:result}
% \end{figure}

\section{Conclusion}
We propose a novel deep learning ensemble model for breast lesion classification, which simultaneously uses digital mammograms (DM) and digital breast tomosynthesis (DBT). We innovatively use the RankSVM algorithm on DBT to extract a fixed representation, dynamic feature image, of DBT. Dynamic feature image captures the slice-to-slice difference in DBT, which is the information often looked by radiologists. The experiments show that when using both DM and DBT in combination, the single model performance can be improved nearly $10\%$ on AUC and $23\%$ on the prediction confidence. By applying ensemble strategy on the three classifiers, the best performance can be improved to $0.97$ AUC. This improvement indicates that deep learning models, like radiologists, benefit from combining both mammographic image formats. Also, the consistency of better performance across different feature extractors and classifiers suggests that our method is not limited to any specific deep learning architecture. The proposed DBT data representation method and dynamic feature image can also increase the classification performance of using DBT-only data by nearly $24\%$. In addition, our approach uses only the image-level labels. Due to a large number of incoming data in the daily clinical practice, annotating images with bounding boxes is not practical. However, we believe that with more precise labels, such as bounding boxes, the performance of our model can be further improved. Our model can adapt to bounding boxes labeling with minor changes.

\bibliographystyle{IEEEtran}
\bibliography{bibfile}

\end{document}